%% file: main.tex
\documentclass[sigconf]{acmart}

\AtBeginDocument{%
  \providecommand\BibTeX{{%
    Bib\TeX}}}

\copyrightyear{2025} 
\acmYear{2025} 
\setcopyright{acmlicensed}\acmConference[MM '25]{Proceedings of the 33rd
ACM International Conference on Multimedia}{October 27--31, 2025}{Dublin,
Ireland}
\acmBooktitle{Proceedings of the 33rd ACM International Conference on
Multimedia (MM '25), October 27--31, 2025, Dublin, Ireland}
\acmDOI{10.1145/3746027.3758161}
\acmISBN{979-8-4007-2035-2/2025/10}

\usepackage{mathtools}      
\usepackage{algorithm}
\usepackage{algorithmic}
\usepackage{soul}
\usepackage{url}
\usepackage{subcaption}     

\begin{document}

\title{Physics-Informed Representation Alignment for Sparse Radio-Map Reconstruction}

\author{Haozhe Jia}
\authornote{Equal contribution.}
\affiliation{
  \institution{HKUST (GZ);}
  \city{Guangzhou}
  \country{China}
}


\author{Wenshuo Chen}
\authornotemark[1]
\affiliation{
  \institution{HKUST (GZ)}
  \city{Guangzhou}
  \country{China}
}

\author{Zhihui Huang}
\authornotemark[1]
\affiliation{
  \institution{Shandong University}
  \city{Qingdao}
  \country{China}
}

\author{Lei Wang}
\affiliation{%
  \institution{Griffith University}
  \city{Brisbane}
  \state{Queensland}
  \country{Australia}
}
\affiliation{%
  \institution{Data61/CSIRO}
  \city{Canberra}
  \state{ACT}
  \country{Australia}
}


\author{Hongru Xiao}
\affiliation{
  \institution{Tongji University}
  \city{Shanghai}
  \country{China}
}

\author{Nanqian Jia}
\affiliation{
  \institution{Peking University}
  \city{Shenzhen}
  \country{China}
}

\author{Keming Wu}
\affiliation{
  \institution{Tsinghua University}
  \city{Beijing}
  \country{China}
}

\author{Songning Lai}
\affiliation{
  \institution{HKUST (GZ)}
  \city{Guangzhou}
  \country{China}
}

\author{Bowen Tian}
\affiliation{%
  \institution{HKUST (GZ)}
  \city{Guangzhou}
  \country{China}
}

\author{Yutao Yue}
\authornote{Correspondence to Yutao Yue <yutaoyue@hkust-gz.edu.cn>}
\affiliation{%
  \institution{HKUST (GZ)}
  \city{Guangzhou}
  \country{China}
}
\affiliation{%
  \institution{Institute of Deep Perception Technology}
  \city{Wuxi}
  \country{China}
}
\settopmatter{authorsperrow=5}

\settopmatter{printacmref=false}
\renewcommand\footnotetextcopyrightpermission[1]{}
\renewcommand{\shortauthors}{Haozhe Jia et al.}

\begin{abstract}
Radio map reconstruction is essential for enabling advanced applications, yet challenges such as complex signal propagation and sparse observational data hinder accurate reconstruction in practical scenarios. Existing methods often fail to align physical constraints with data-driven features, particularly under sparse measurement conditions. To address these issues, we propose \textbf{Phy}sics-Aligned \textbf{R}adio \textbf{M}ap \textbf{D}iffusion \textbf{M}odel (\textbf{PhyRMDM}), a novel framework that establishes cross-domain representation alignment between physical principles and neural network features through dual learning pathways. The proposed model integrates \textbf{Physics-Informed Neural Networks (PINNs)} with a \textbf{representation alignment mechanism} that explicitly enforces consistency between Helmholtz equation constraints and environmental propagation patterns. Experimental results demonstrate significant improvements over state-of-the-art methods, achieving \textbf{NMSE of 0.0031} under Static Radio Map (SRM) conditions, and \textbf{NMSE of 0.0047} with in Dynamic Radio Map (DRM) scenarios. The proposed representation alignment paradigm provides 37.2\% accuracy enhancement in ultra-sparse cases (1\% sampling rate), confirming its effectiveness in bridging physics-based modeling and deep learning for radio map reconstruction. The code can be found on the website: \textcolor{blue}{\href{https://github.com/Hxxxz0/RMDM}{Code}}
\end{abstract}




\maketitle

\input{sec/1_intro}

\input{sec/2_Relatedwork}

\input{sec/method}

\section{Experiment}
\input{sec/Experiment}


\input{sec/conclusion}

\bibliographystyle{ACM-Reference-Format}
\bibliography{acmart}


\end{document}

%% file: sec/1_intro.tex
\section{Introduction}

With the continuous breakthroughs made by humans in information technology, some new concepts and technologies have begun to emerge, such as 5G, 6G communications, terahertz communications, and autonomous driving technology \cite{10579941,9493470,9046805}. In order to fully utilize the advantages of various technologies, It is particularly important to plan the base station layout and allocate frequency resources precisely and efficiently \cite{Value_of_Frequency_Spectrum}.

\begin{figure}[t]
    \centering
    \includegraphics[width=\linewidth]{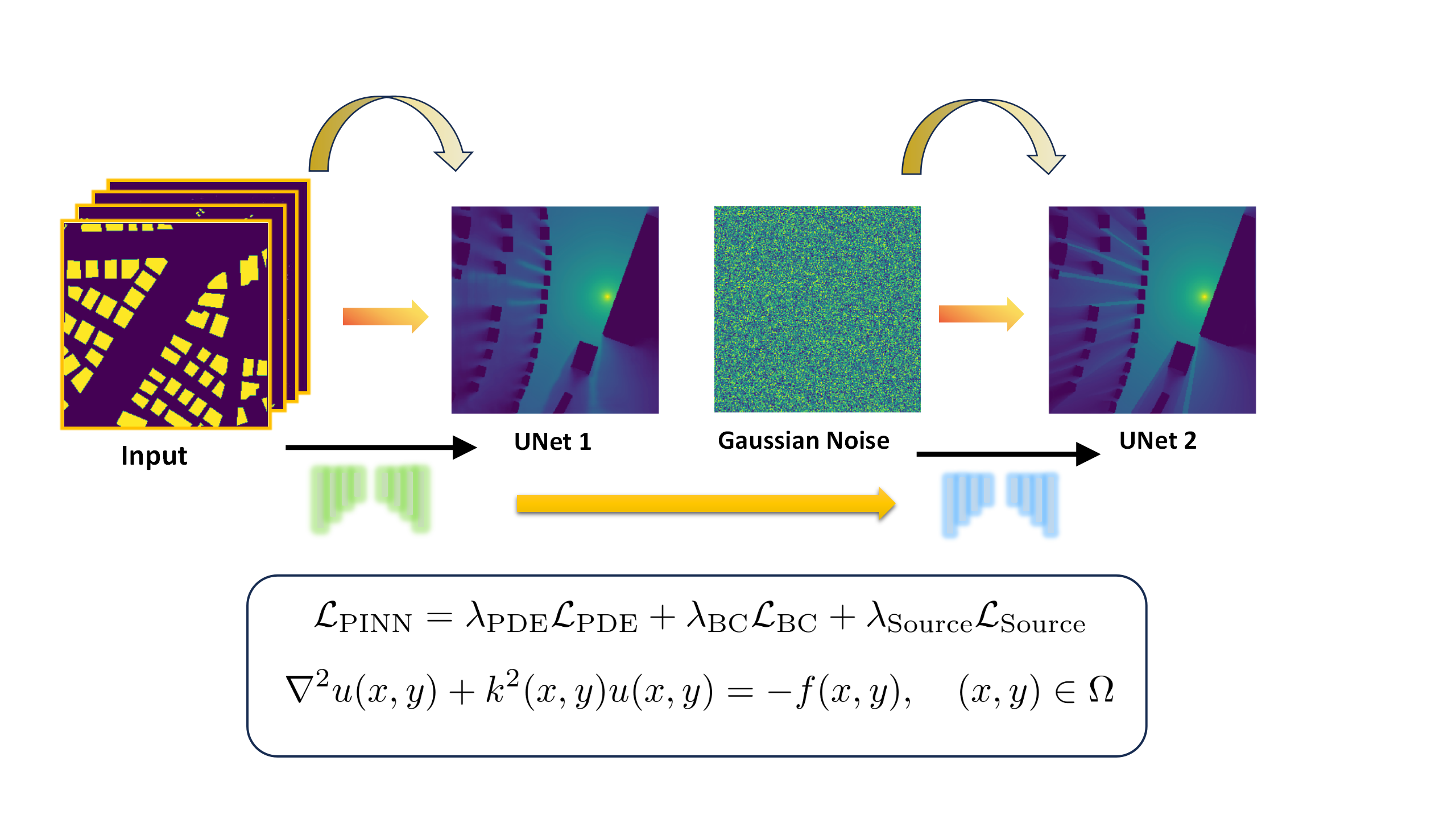} 
    \caption{Schematic of the PhyRMDM architecture integrating the Helmholtz equation for path loss. The dual UNet design uses UNet 1 with PINN-based physical constraints and UNet 2 for diffusion-based refinement. Guided by losses ($L_\text{PDE}$, $L_\text{BC}$, $L_\text{Source}$), it ensures accurate and robust radio map predictions.} 
    \label{fig:intro}
\end{figure}

Planning and allocation are closely linked to radio frequency (RF) radio maps. These maps are like a spatial snapshot of RF signals. They are really important because they show how signal power is spread out in different areas. This distribution, referred to as the geographical signal power spectral density (PSD), quantifies the accumulation of RF signal power as a function of spatial location and frequency. It also takes into account how these factors are related and how they might change over time. Geographical characteristics influence signal paths \cite{GPSD} based on RF propagation properties, with signal strength quantified in dBm and associated with power attenuation. Two principal mapping approaches \cite{RF-filed-measurement} exist: field measurement, which uses professional RF equipment for precise power capture and attenuation assessment but yields localized, discontinuous data \cite{RF-filed-measurement} and model prediction, such as ray-tracing methods, which employ mathematical and physical principles to analyze propagation environments comprehensively, enabling large-scale data generation without extensive on-site measurements \cite{frei2009prediction}. However, manual collection methods have disadvantages such as high cost and low efficiency.

Most current radiomap estimation approaches fall into two main categories: model-based and model-free methods, such as REM-GAN \cite{REM-GAN} and RadioGAT \cite{RadioGAT}. Model-based methods rely on predefined signal propagation models, like the log-distance path loss model for Wi-Fi radio map reconstruction or interpolation based on thin-plate spline kernels \cite{Interpolation1}, but they struggle to capture complex environmental effects such as shadowing and obstacles. Model-free methods, on the other hand, explore neighborhood information without assuming specific models, employing techniques such as Radial Basis Function interpolation and inverse distance weighting. While more flexible, they depend heavily on the quality of observed samples and often assume uniform data distributions—an assumption that fails in practical scenarios like MDT, where user measurements are unevenly distributed. Additionally, variations in propagation models and parameters across training datasets further complicate the reconstruction process. As a result, integrating model-based and model-free methods offers a promising avenue for overcoming these challenges effectively \cite{MBMF1l}.

Diffusion models \cite{diff1-nichol2021improved,song2020denoising} have emerged as a robust framework in generative modeling \cite{sato,fts}, achieving state-of-the-art results in image synthesis \cite{rombach2022highresolutionimagesynthesislatent,ramesh2021zeroshottexttoimagegeneration, ning2024dctdiffintriguingpropertiesimage}, inpainting \cite{yang2023diffusion} and other tasks \cite{hu2024multidimensionalexplanationalignmentmedical, song2022denoisingdiffusionimplicitmodels}. These models leverage iterative denoising to effectively model complex data distributions, enabling the generation of high-quality \cite{ho2020denoisingdiffusionprobabilisticmodels,ning2024dctdiffintriguingpropertiesimage}, photorealistic images. Their flexibility and mathematical rigor have facilitated advancements in diverse applications, including creative content generation and medical imaging \cite{croitoru2023diffusion}. Recently, diffusion models have been applied to radio map reconstruction, where capturing the spatial distribution of signal power is critical. For instance, RadioDiff \cite{wang2024radiodiff} demonstrates how these models address the challenges posed by irregular and sparse signal environments by simulating the propagation of radio waves \cite{sizun2005introduction}. However, the limitation of data volume is a major challenge faced by the entire model field \cite{l2017machine}. It severely restricts the model's ability to learn the laws of the complex physical world, and the diffusion model is no exception. After in-depth analysis, We decided to solve this problem by embedding physical laws to achieve the representational alignment of physical information.

\textbf{Just as humans use physical laws to better predict natural phenomena, do models benefit in the same way?} Even physical phenomena that have not occurred or been observed can be predicted. To leverage this capability, we combined our model with the Physics-Informed Neural Network (PINN) \cite{lawal2022physics}, integrating physical constraints to enhance performance and improve generalization ability, surpassing traditional data-driven approaches. Unlike conventional neural networks that rely heavily on extensive datasets to identify patterns, this way incorporate physical principles, enabling the model to utilize prior knowledge and physical constraints. This hybrid approach effectively addresses challenges such as limited data availability and complex propagation environments. During PINN training \cite{monaco2023training,nabian2021efficient}, the optimization process minimizes not only the traditional loss of data fit-representing not only the deviation between the predicted and observed values but also the residuals of the embedded physical equations, ensuring that the model is in line with the representation of the physical information while improving the accuracy and robustness of the prediction.

However, directly applying physical constraints to diffusion models is challenging. Real-world radio environments are too complex to be described by a single, strict equation, and diffusion models predict noise, making direct physical loss application impractical. To address this, we designed a dual U-Net architecture. One U-Net (U-Net 1) is constrained by an approximate two-dimensional Helmholtz equation to provide physical representation guidance. In parallel, a second U-Net (U-Net 2) \cite{ronneberger2015u} refines these features through a diffusion process, generating high-resolution radio maps. By leveraging the \textbf{Helmholtz} equation to inform the learning process, we enhance the model's ability to generate accurate predictions and improve its generalization in scenarios with sparse or unevenly distributed data.

This physical information representation alignment and machine learning principles enables PhyRMDM to accurately capture complex spatial signal characteristics \cite{RAISSI2019686}. In experiments conducted under the Static RM (SRM) setting, our model achieved an \textbf{NMSE of 0.0031} and an \textbf{RMSE of 0.0125}, outperforming state-of-the-art (SOTA) approaches. Extensive experimental validation further highlights the model's superior performance in reconstructing radio maps across various challenging scenarios, establishing a novel paradigm that combines physics-driven methodologies with data-driven techniques to address real-world signal propagation challenges \cite{LI20203394}.

\begin{itemize}
    \item We propose the PhyRMDM, a physics-informed representation consistency framework that integrates the \textbf{Helmholtz equation} and diffusion processes within a dual U-Net architecture. This design ensures physical consistency and significantly enhances the accuracy of radio map reconstruction in sparse and complex propagation scenarios.

    \item By aligning representation of Physics-Informed Neural Networks (PINNs), PhyRMDM embeds physical information representation alignment into the learning process, enabling precise modeling of spatial signal characteristics, such as signal strength and path loss, while improving generalization across diverse environments.

    \item Extensive experimental validation demonstrates that PhyRMDM achieves \textbf{state-of-the-art} performance, including an \textbf{NMSE of 0.0031} and \textbf{RMSE of 0.0125} under the Static RM (SRM) setting, highlighting its robustness and scalability for real-world signal propagation challenges.
\end{itemize}

%% file: sec/2_Relatedwork.tex
\section{Related Work}

\subsection{Radio Map Estimation}

Radio map estimation is essential for applications such as network planning and spectrum management. Traditional methods include model-based approaches, which utilize established signal propagation models like the log-distance path loss (LDPL) model for Wi-Fi radio map reconstruction \cite{jung2011wi}. Interpolation techniques, such as thin-plate splines \cite{keller2019thin}, have also been employed to achieve spatially continuous estimations. However, these methods often assume ideal conditions and may not accurately capture complex environmental factors like shadowing and multipath effects, leading to limitations in heterogeneous settings.

\subsection{Diffusion Models}
Diffusion models can capture complex data distributions and have been applied to radio map estimation. RadioDiff~\cite{diff1-nichol2021improved} enables sampling-free dynamic radio map construction, while RM-Gen~\cite{luo2024rm} uses conditional denoising diffusion to synthesize maps from minimal data. These methods reduce reliance on extensive measurements and costly simulations.

\subsection{Physics-Informed Neural Networks (PINNs)}

Physics-Informed Neural Networks (PINNs)~\cite{raissi2019physics} incorporate physical laws, typically expressed as partial differential equations (PDEs)~\cite{sloan2012partial}, into neural network training to ensure predictions comply with known physics, improving interpretability and generalization in sparsedata scenarios. For radio map estimation, embedding the Helmholtz equation—describing wave propagation—into the loss function enhances accuracy and robustness under challenging measurement conditions.

PINNs have been applied across diverse scientific and engineering domains~\cite{hu2023applying,eghtesad2024nn}, such as solving the Navier–Stokes equations in fluid mechanics, incorporating constitutive laws in materials science, improving climate forecasts via conservation laws, and modeling electromagnetic wave propagation. These applications demonstrate that physical constraints reduce overfitting and boost interpretability, especially with limited or noisy data.

Unlike conventional PINNs that enforce PDE residuals only at the output, we adopt a \emph{Representation Alignment} strategy that aligns latent features with high-quality priors, ensuring both physical consistency and semantic discriminability.

%% file: sec/method.tex
\section{Method}

\begin{figure*}[t!] 
  \centering 
  \includegraphics[width=0.9\textwidth]{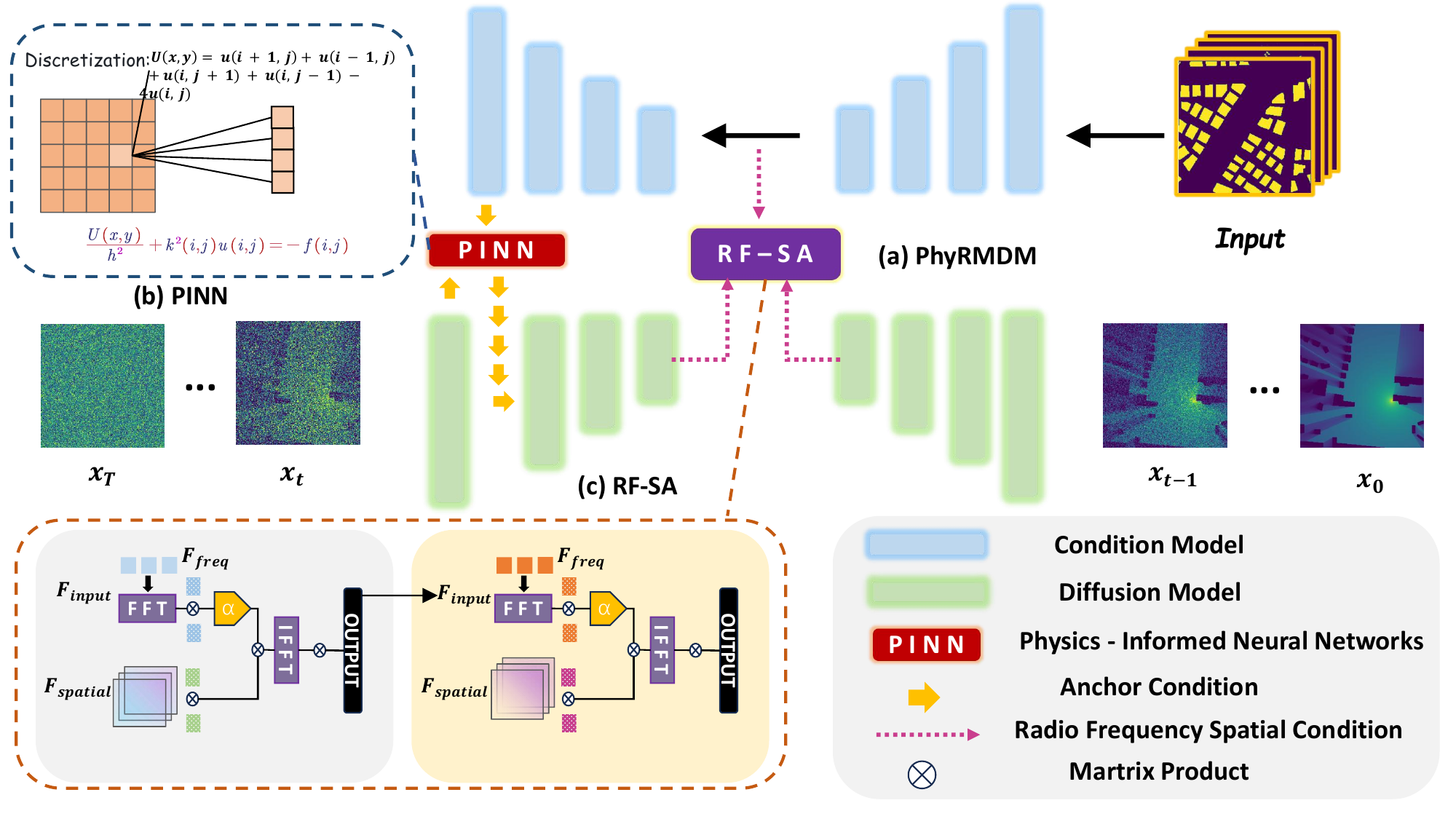}
  \vspace{1em}
  \caption{An illustration of the PhyRMDM framework. (a) shows the pipeline overview, combining a condition model and a diffusion model for radio map reconstruction. Key components include (b) a physics-constrained feature alignment module, which integrates physical priors with learned representations, and (c) Radio Frequency Spatial Attention (RF-SA), which aligns spatial and frequency features to enhance accuracy. The framework leverages anchor conditions and representation alignment to iteratively refine features, ensuring accurate predictions in complex propagation scenarios.}
  \label{fig:main} 
\end{figure*}

The radio map estimation task aims to predict the spatial distribution of wireless signal strength within a given area. Wireless signal propagation follows physical laws, such as electromagnetic wave equations and path loss models \cite{jiang2024physics}. Incorporating these physical constraints into learning-based models can significantly improve the accuracy and reliability of predictions. Here is a complementary approach we have developed : aligning learned representations with domain-specific priors to improve model efficiency and robustness.

As illustrated in Figure \ref{fig:main}, we propose a diffusion-based probabilistic framework that integrates a dual UNet architecture with lightweight physical constraint modules. The framework emphasizes \textbf{representation alignment} between multi-scale features and domain-specific signal characteristics, structured in two stages:

\begin{itemize}
    \item \textbf{Conditional Generation Model (UNet 1):}In this stage, input features are extracted and a Physics-Informed Neural Network (PINN) module enforces physical constraints to achieve physical information representational alignment.
    \item \textbf{Diffusion Model (UNet 2):} In this stage, features are refined through a diffusion process to generate the final radio map estimation.
\end{itemize}

\subsection{Problem Formulation}
Reconstructing the radio map involves modeling the spatial distribution of wireless signals, which can be formulated using the Helmholtz equation \cite{juraev2024helmholtz}. This equation provides a physical foundation for capturing signal propagation dynamics:

\begin{equation}
\nabla^2 u(x, y) + k^2(x, y) u(x, y) = -f(x, y), \quad (x, y) \in \Omega.
\end{equation}

\noindent where $u(x, y)$ is the wireless signal strength, $\nabla^2$ is the Laplacian operator, $k(x, y)$ is the wavenumber accounting for signal attenuation and dispersion, $f(x, y)$ is the source term, and $\Omega$ is the domain of interest.

In a discrete computational framework, the Laplacian $\nabla^2 u(x, y)$ is approximated using the central difference method, resulting in the discretized form:

\begin{equation}
\small
\begin{aligned}
\Delta_h u(i, j) &= \frac{u(i + 1, j) + u(i - 1, j) + u(i, j + 1) + u(i, j - 1)}{h^2} \\
& \quad - \frac{4u(i, j)}{h^2},
\end{aligned}
\end{equation}

\noindent where $h$ denotes the grid spacing. Substituting this into the Helmholtz equation, we define the residual at grid point $(i, j)$ as:
\begin{equation}
r(i, j) = \Delta_h u(i, j) + k^2(i, j) u(i, j) - f(i, j),
\end{equation}

\noindent where $r(i, j)$ represents the discrepancy between the left and right-hand sides of the discrete equation.

To enforce the physical consistency of the network output, we define a PDE-constrained loss function:
\begin{equation}
\mathcal{L}_{\text{PDE}} = \frac{1}{N_{\text{int}}} \sum_{(i, j) \in \Omega_{\text{int}}} [r(i, j)]^2,
\end{equation}

\noindent where $N_{\text{int}}$ is the number of interior grid points.

Dirichlet boundary conditions, defined as $u(i, j) = u_{\text{BC}}(i, j)$, such as those encountered in applications involving buildings or vehicles, are enforced using a boundary loss term:
\begin{equation}
\mathcal{L}_{\text{BC}} = \frac{1}{N_{\text{bc}}} \sum_{(i, j) \in \partial \Omega} [u(i, j) - u_{\text{BC}}(i, j)]^2,
\end{equation}

\noindent where $\partial \Omega$ represents the set of boundary grid points, and $u_{\text{BC}}(i, j)$ denotes the prescribed boundary values.

Additionally, if source data is provided, a source loss is defined as:
\begin{equation}
\mathcal{L}_{\text{Source}} = \frac{1}{N_{\text{src}}} \sum_{(i, j) \in \Omega_{\text{src}}} [u(i, j) - u_{\text{src}}(i, j)]^2,
\end{equation}

\noindent where $\Omega_{\text{src}}$ includes the grid points with known source values.

The total loss function combines these terms as:
\begin{equation}
\mathcal{L}_{\text{PINN}} = \lambda_{\text{PDE}} \mathcal{L}_{\text{PDE}} + \lambda_{\text{BC}} \mathcal{L}_{\text{BC}} + \lambda_{\text{Source}} \mathcal{L}_{\text{Source}},
\end{equation}

\noindent where $\lambda_{\text{PDE}}$, $\lambda_{\text{BC}}$, and $\lambda_{\text{Source}}$ are weighting factors that balance the contributions of each term.

This formulation ensures that the network output adheres to the physical principles governing the propagation of the wireless signal while effectively incorporating the boundary and source constraints.

\subsection{Model Architecture}
Directly applying physical constraints to diffusion models is challenging, as they predict noise rather than signal strength, and complex radio environments are difficult to model with a single equation. To address this, our model employs a dual U-Net architecture. One U-Net (U-Net 1) is constrained by an approximate two-dimensional Helmholtz equation to provide physical representation guidance. In parallel, a second U-Net (U-Net 2) refines these features through a diffusion process to generate the final radio map \cite{kawar2022denoising}. To enhance feature transfer between the two U-Nets, we propose two innovative mechanisms: the Anchor Condition and RF-SA.

\subsubsection{UNet 1: Conditional Generation Model with PINN Alignment}
UNet 1 serves as the initial stage for feature extraction and embedding physics-informed constraints. The input consists of observed radio signal data $x \in \mathbb{R}^{C_{\text{obs}} \times H \times W}$, where $C_{\text{obs}}$ represents signal channels (e.g., signal strength, path loss), and a spatial mask $m \in \{0, 1\}^{H \times W}$ indicating observation locations.

The network extracts multi-scale features $\{z_1, z_2, \dots, z_L\}$ and incorporates a representation alignment module in the decoder, which ensure adherence to physical laws while prioritizing alignment with pre-trained radio signal representations.

The loss function is defined as:
\begin{equation}
\mathcal{L}_{\text{Cond}} = \lambda_{\text{MSE}} \mathcal{L}_{\text{MSE}} + \lambda_{\text{PINN}} \mathcal{L}_{\text{PINN}} + \lambda_{\text{Reg}} \mathcal{L}_{\text{Reg}},
\end{equation}

\noindent where $\mathcal{L}_{\text{MSE}} = \frac1N\sum_i (\hat{y}_i - y_i)^2$ensures the effectiveness and accuracy of feature extraction by aligning the model output with the real data. $\mathcal{L}_{\text{Reg}}$ represents a regularization term that promotes smoothness or penalizes overfitting, depending on the specific design of the model or the application.

\subsubsection{UNet 2: Diffusion Model for Feature Refinement and Map Generation}
UNet 2 serves as the second and critical stage in the overall framework, focusing on refining the features extracted by UNet 1 and generating the final radio map estimation. Its input is the output $\hat{y}$ from UNet 1, which incorporates initial physics-informed features and basic signal characteristics.

The diffusion process in UNet 2 begins with Gaussian noise, $x_T \sim \mathcal{N}(0, I)$. Through an iterative denoising process, the model progressively refines the input features to approximate the true spatial distribution of radio signals. To enhance the integration of the conditional features from UNet 1, an Anchor Condition mechanism is introduced. This mechanism is expressed as:
\begin{equation}
f_t = \sigma(f_{\text{d,\,t}}) \cdot \phi(f_{\text{c,\,t}})
\end{equation}

\noindent where $f_{\text{d,\,t}}$ and $f_{\text{c,\,t}}$ represent the diffusion-related and condition-related features, respectively, while $\sigma$ and $\phi$ are implemented using $1\times1$ convolutions and activation functions.

The loss function for UNet 2 is defined as the diffusion loss:
\begin{equation}
\mathcal{L}_{\text{Diff}} = \mathbb{E}_{x_0, t, \epsilon} [\|\epsilon - \epsilon_{\theta}(x_t, t, \hat{y})\|^2],
\end{equation}

\noindent where $\epsilon$ denotes the noise to be predicted. This loss function plays a critical role in guiding the model to optimize the denoising process during diffusion, enabling it to learn the underlying signal distribution effectively.

By leveraging the pre-processed features from UNet 1 and refining them through the diffusion mechanism, UNet 2 achieves enhanced accuracy and reliability in the final radio map estimation. This architecture ensures the model fully capitalizes on both physics-informed priors and iterative refinement to produce high-quality predictions.

\subsubsection{Overview of Model Architecture}
The proposed framework consists of two main components: UNet 1 for feature extraction and representation alignment and UNet 2 for feature refinement and radio map generation. The overall training objective is defined by the total loss function:
\begin{equation}
\mathcal{L}_{\text{Total}} = \lambda_{\text{Cond}} \mathcal{L}_{\text{Cond}} + \lambda_{\text{Diff}} \mathcal{L}_{\text{Diff}},
\end{equation}

\noindent where $\lambda_{\text{Cond}}$ and $\lambda_{\text{Diff}}$ are weighting factors that balance the contributions of each stage.

In addition to the Anchor Condition mechanism, we introduce RF-SA (Radio Frequency-Spatial Attention) to enhance feature transfer between the two U-Net components. RF-SA leverages spectral attributes by transforming features into the frequency domain using the Fast Fourier Transform (FFT), integrating them with spatial features through an attention mechanism:

\begin{equation}
F_{\text{attn}} = \alpha(\text{FFT}(F_{\text{input}})) \cdot F_{\text{spatial}},
\end{equation}

\noindent where $\alpha$ represents a non-linear activation. This mechanism addresses challenges in signal propagation and interference, ensuring effective feature fusion and enhancing radio map reconstruction.

\input{sec/algorithm}

%% file: sec/algorithm.tex
\begin{algorithm}[tb]
\caption{Physics-Aligned Radio Map Diffusion Model (Phy-RMDM)}
\label{alg:rmdm}
\textbf{Input}: Radio data $x \in \mathbb{R}^{C \times H \times W}$, mask $m \in \{0,1\}^{H \times W}$, noise levels $T$.\\
\textbf{Parameter}: weights~$\lambda_{\text{MSE}}, \lambda_{\text{Reg}}, \lambda_{\text{PDE}}, \lambda_{\text{BC}}, \lambda_{\text{Source}}, \lambda_{\text{Cond}}, \lambda_{\text{Diff}}$.\\
\textbf{Output}: Reconstructed radio map $\hat{y}$.

\begin{algorithmic}[1]
\STATE Initialize U-Net$1$($\theta_1$), U-Net$2$($\theta_2$)
\WHILE{not converged}
    \STATE \textbf{Stage 1: Physics-Guided Feature Extraction}
    \STATE Generate initial estimate $\hat{y}_0 = \text{U-Net}^1_\theta(x \odot m)$
    \STATE Extract multi-scale features $\{z_l\}$ from U-Net$^1$ encoder
    \STATE Compute physics losses: 
    \STATE $\mathcal{L}_{\text{PINN}} = \lambda_{\text{PDE}}\mathcal{L}_{\text{PDE}} + \lambda_{\text{BC}}\mathcal{L}_{\text{BC}} + \lambda_{\text{Source}}\mathcal{L}_{\text{Source}}$
    \STATE Compute $\mathcal{L}_{\text{Cond}} = \lambda_{\text{MSE}}\|\hat{y}_0-y\|^2 + \mathcal{L}_{\text{PINN}} + \lambda_{\text{Reg}}\mathcal{L}_{\text{Reg}}$
    
    \STATE \textbf{Stage 2: Conditional Diffusion Process}
    \STATE Sample $t \sim \mathcal{U}(1,T)$, $\epsilon \sim \mathcal{N}(0,I)$
    \STATE Corrupt estimate: $\hat{y}_t = \sqrt{\alpha_t}\hat{y}_0 + \sqrt{1-\alpha_t}\epsilon$
    \STATE Apply RF-SA: $z'_l = \text{RF-SA}(z_l, x_{\text{RF}})$ \COMMENT{Radio Frequency-Spatial Alignment}
    \STATE Predict noise: $\epsilon_\theta = \text{U-Net}^2_\theta(\hat{y}_t, t, \{\text{Anchor}(z'_l)\})$ 
    \STATE Compute $\mathcal{L}_{\text{Diff}} = \|\epsilon - \epsilon_\theta\|^2$
    
    \STATE \textbf{Update Parameters}
    \STATE $\mathcal{L}_{\text{Total}} = \lambda_{\text{Cond}}\mathcal{L}_{\text{Cond}} + \lambda_{\text{Diff}}\mathcal{L}_{\text{Diff}}$
    \STATE Update $\theta_1, \theta_2$ via $\nabla_{\theta}\mathcal{L}_{\text{Total}}$
\ENDWHILE
\STATE Generate final $\hat{y}$ via diffusion ancestral sampling conditioned on $\hat{y}_0$
\end{algorithmic}
\end{algorithm}

%% file: sec/Experiment.tex
\begin{figure}[t!]
    \centering
    \includegraphics[width=\linewidth]{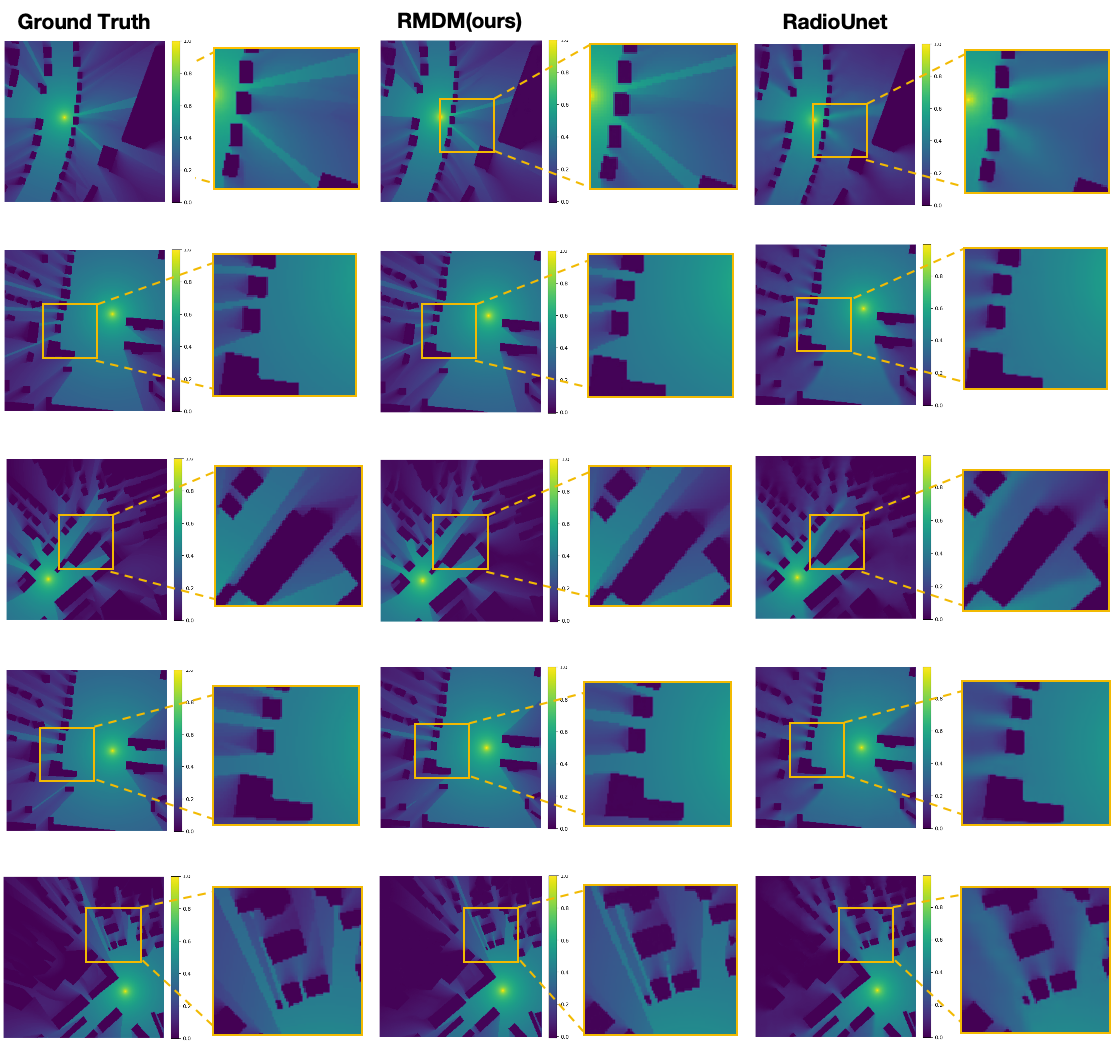} 
    \caption{Visualization of the results generated by PhyRMDM compared to RadioUnet. The proposed PhyRMDM model demonstrates more fine-grained results and exhibits closer alignment with the ground truth data, highlighting its superiority in capturing intricate details.} 
    \label{fig:Visualization}
\end{figure}

This study evaluates PhyRMDM, a framework designed to overcome the inherent challenges of applying physical constraints to diffusion-based generative models for radio map reconstruction. Using the RadioMapSeer dataset from the Pathloss RM Construction Challenge, we validate our approach across three distinct experimental setups, comparing its performance against state-of-the-art models like RME-GAN, RadioUNet, and RadioDiff. The results demonstrate that our model consistently outperforms these methods. We also conduct a detailed ablation study to assess the contribution of individual components. As shown in Figure~\ref{fig:Visualization}, we provide a visual comparison with RadioUNet, highlighting the improved quality and precision of our approach.

\subsection{Dataset}

The RadioMapSeer dataset \cite{rms} contains 700 maps with unique geographic data, including 50–150 buildings and 80 transmitter locations per map, plus ground truth. We use 500 maps for training and 200 for testing with no terrain overlap.

Maps from OpenStreetMap cover cities such as Ankara, Berlin, Glasgow, Ljubljana, London, and Tel Aviv. Transmitter, receiver, and building heights are set to 1.5 m, 1.5 m, and 25 m, respectively, with transmitter power at 23 dBm and carrier frequency at 5.9 GHz. Each map is a 256×256 binary image (1 m/pixel): ‘1’ for building interiors, ‘0’ otherwise; transmitter positions are similarly encoded with the transmitter pixel set to ‘1’.

The radio maps in this dataset are generated using software such as WinProp \cite{jakobus2018recent}. To ensure accurate ground truth for training, the dataset employs Maxwell’s equations to construct radio maps (RMs). Pathloss is calculated by considering the reflection and diffraction of electromagnetic rays. The dataset includes two types of RMs: Static RM (SRM), which considers the impact of static buildings, and Dynamic RM (DRM), which includes both static buildings and randomly generated vehicles along the roads. This comprehensive data set enables a robust evaluation of the performance of our model in realistic urban conditions.

\subsection{Experiment Setup}

We conducted experiments under the following three settings to evaluate the performance of our model:

\begin{itemize}
    \item \textbf{Setup 1 (SRM)}: The input includes building information and transmitter location data.
    \item \textbf{Setup 2 (DRM)}: The input includes building information, transmitter location data, and vehicle information.
    \item \textbf{Setup 3 (Unbalanced Sample Distribution Among Regions)}: In this setup, we sample the radio map at a random ratio between 1\% and 10\% in each region. The sparse observations are uniformly distributed across the sampled areas.
\end{itemize}

The training was conducted on an NVIDIA RTX 4090 GPU, with the model trained for approximately 500,000 steps to ensure convergence and optimal performance. The hyperparameters for the loss function were carefully tuned to balance different objectives. Specifically, the weights assigned to the various components were as follows: $\lambda_{\text{MSE}}$ (1.0), $\lambda_{\text{Reg}}$ (0.1), $\lambda_{\text{PDE}}$ (1.5), $\lambda_{\text{BC}}$ (0.5), $\lambda_{\text{Source}}$ (0.2), $\lambda_{\text{Cond}}$ (1.0), and $\lambda_{\text{Diff}}$ (1.0). These hyperparameters were meticulously tuned to achieve an optimal trade-off between accuracy, generalization, and adherence to the underlying physical principles.

\begin{figure*}[ht]
    \centering
    \includegraphics[width=1\linewidth]{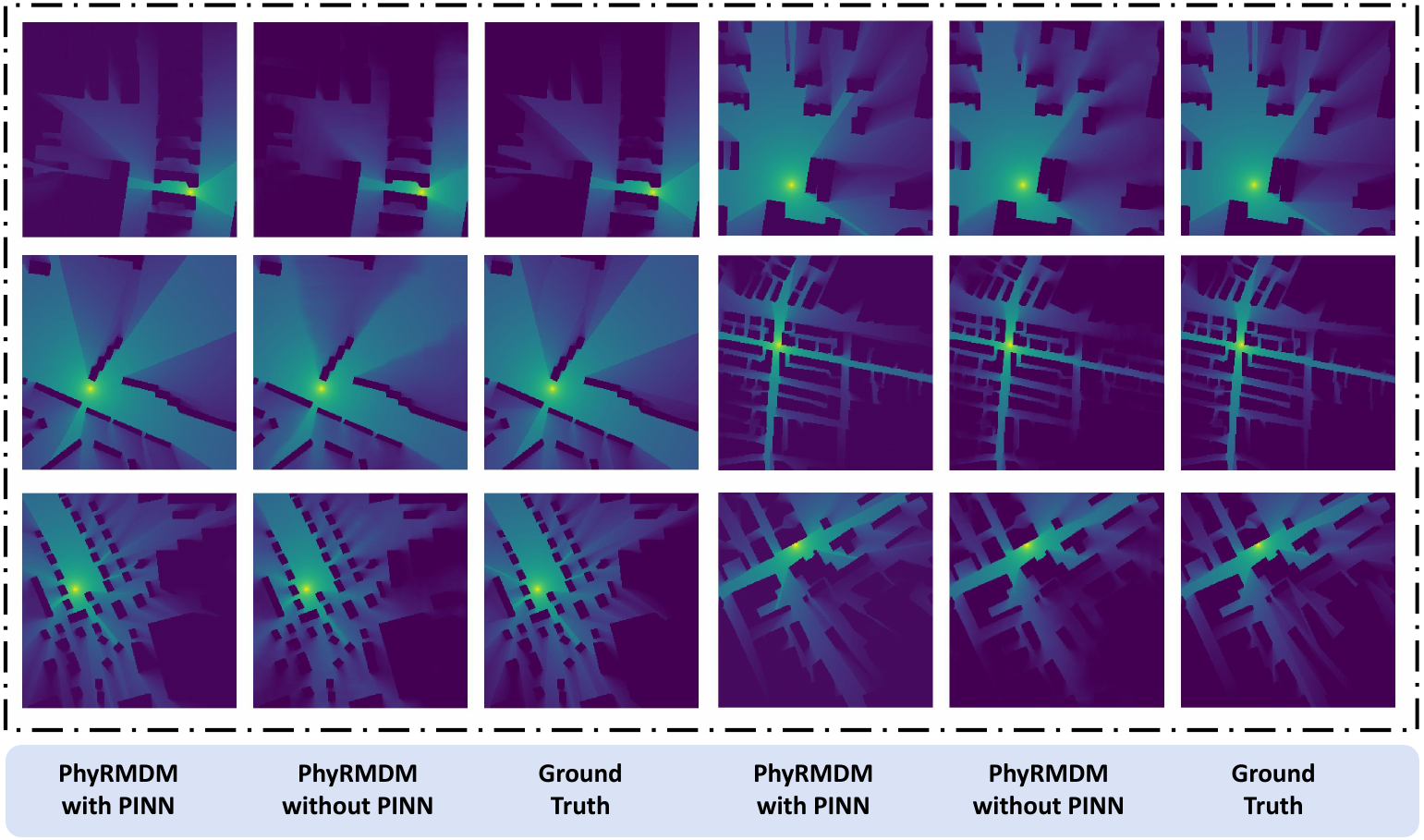}
    \caption{Visual comparison of ablation study on sparse radio map reconstruction. Each triplet shows results from PhyRMDM with PINN, without PINN, and ground truth. PINN enhances structural details and accuracy, highlighting its effectiveness in physics-guided reconstruction.}
    \label{fig:ablation_PINN}
\end{figure*}

\subsection{Results and Analysis}

To evaluate the performance of our proposed model, we conducted experiments under three distinct setups: (1) Static Radio Maps (SRM), (2) Dynamic Radio Maps (DRM), and (3) Unbalanced Sample Distribution Among Regions. The results for each setup are detailed below, demonstrating the efficacy of PhyRMDM in comparison to existing state-of-the-art methods.
\input{sec/tab/SRM}
\subsubsection{Setup 1: Static Radio Maps (SRM)}

In this configuration, we utilized static building information and transmitter location data as inputs to assess the model's performance. As presented in Table \ref{tab:setup1}, our proposed model, PhyRMDM, outperforms existing models across all metrics. Specifically, PhyRMDM achieves the lowest NMSE of\textbf{ 0.0031} and RMSE of \textbf{0.0125}, while attaining the highest SSIM of \textbf{0.978}. These results indicate that PhyRMDM surpasses state-of-the-art models such as RadioDiff, RadioUNet, and RME-GAN, demonstrating its capability to generate precise and structurally consistent radio maps in static scenarios.

\subsubsection{Setup 2: Dynamic Radio Maps (DRM)}

To evaluate the model's performance under dynamic conditions, we incorporated vehicle information into the input data alongside building and transmitter location information. The results in Table \ref{tab:setup2} illustrate that PhyRMDM consistently outperforms competing models, achieving an \textbf{NMSE} of \textbf{0.0047}, an \textbf{RMSE} of \textbf{0.0146}, and an \textbf{SSIM} of \textbf{0.968}. These metrics demonstrate the robustness of PhyRMDM in modeling dynamic urban environments, where additional complexity is introduced by moving objects such as vehicles.
\input{sec/tab/DRM}
\subsubsection{Setup 3: Unbalanced Sample Distribution}

In the third setup, we tested the model's ability to handle unbalanced sample distributions, with sampling ratios varying randomly between 1\% and 10\%. Table \ref{tab:setup3} shows that PhyRMDM achieves the best performance among all methods, with an \textbf{NMSE} of \textbf{0.0022} and an \textbf{RMSE} of \textbf{0.0117}. PhyRMDM significantly outperforms traditional interpolation-based methods, such as Kriging and RBF, as well as deep learning approaches like AE, Deep AE, and Unet. These results highlight PhyRMDM's capability to generate high-quality radio maps even in the presence of sparse and unevenly distributed observations.

\subsubsection{Overall Performance}

Across all setups, PhyRMDM outperforms state-of-the-art methods, achieving lower NMSE and RMSE with higher SSIM. These results highlight its robustness, generalizability, and efficiency, demonstrating its effectiveness for accurate and reliable radio map construction.

\subsection{Ablation Study}

To investigate the contribution of each loss component ($\mathcal{L}_\text{MSE}$, $\mathcal{L}_\text{PINN}$, $\mathcal{L}_\text{Reg}$) to the overall performance of the model, we conducted a series of ablation experiments, as summarized in Table \ref{tab:ablation_study}. The results indicate that the combination of $\mathcal{L}_\text{MSE}$ and $\mathcal{L}_\text{PINN}$ achieves relatively low errors, with an NMSE of 0.0041 and an RMSE of 0.0155, suggesting that the physics-informed constraints imposed by $\mathcal{L}_\text{PINN}$ are effective in improving prediction accuracy and maintaining physical consistency. However, excluding $\mathcal{L}_\text{MSE}$ while retaining $\mathcal{L}_\text{PINN}$ and $\mathcal{L}_\text{Reg}$ results in a dramatic increase in both NMSE and RMSE, reaching 0.3716 and 0.2413, respectively. This outcome underscores the essential role of $\mathcal{L}_\text{MSE}$ in minimizing overall prediction errors and aligning the output with the ground truth.

In contrast, the model utilizing $\mathcal{L}_\text{MSE}$ and $\mathcal{L}_\text{Reg}$ without $\mathcal{L}_\text{PINN}$ achieves an NMSE of 0.0046 and an RMSE of 0.0182, demonstrating that regularization can partially enhance stability but does not fully capture the underlying physical constraints. The complete model, which incorporates all three loss components, achieves the best performance, with the lowest NMSE of 0.0031 and RMSE of 0.0125. This significant improvement highlights the synergistic effect of combining $\mathcal{L}_\text{MSE}$ for error minimization, $\mathcal{L}_\text{PINN}$ for embedding physics-based priors, and $\mathcal{L}_\text{Reg}$ for ensuring robustness and generalization. 

\input{sec/tab/setup3}

\input{sec/tab/ablation}

Figure~\ref{fig:ablation_PINN} compares results with and without $\mathcal{L}\text{PINN}$. Without it, models produce blurred boundaries and oversmoothed transitions near buildings, failing to capture diffraction and reflection due to missing Helmholtz-based guidance. Adding $\mathcal{L}\text{PINN}$ enhances structural fidelity and enables fine-grained, physically consistent predictions even with sparse data, highlighting its importance for realistic signal modeling.

Overall, combining all loss components yields the best accuracy and stability, balancing error reduction, physical consistency, and regularization for reliable radio map reconstruction.


\vspace{-0.3cm}

%% file: sec/tab/SRM.tex
\begin{table}[t]
    \centering
    \begin{tabular}{lccc}
        \toprule
        Model & NMSE $\downarrow$ & RMSE $\downarrow$ & SSIM $\uparrow$ \\
        \midrule
        RME-GAN & 0.0115 & 0.0303 & 0.9323 \\
        RadioUNet & 0.0074 & 0.0244 & 0.9592 \\
        UVM-Net & 0.0085 & 0.0304 & 0.9320 \\
        RadioDiff & 0.0049 & 0.0190 & 0.9691 \\ \midrule
        \textbf{PhyRMDM (ours)} & \textbf{0.0031} & \textbf{0.0125} & \textbf{0.978} \\ 
        \bottomrule
    \end{tabular}
    \caption{Performance comparison of different models on the SRM setup in terms of NMSE, RMSE, and SSIM. PhyRMDM achieves the lowest NMSE/RMSE and highest SSIM, demonstrating its ability to generate accurate and structurally consistent radio maps. These results confirm its effectiveness in combining physical constraints with diffusion-based modeling for static propagation environments.}
    \label{tab:setup1}
\end{table}

%% file: sec/tab/DRM.tex
\begin{table}[t]
    \centering
    \begin{tabular}{lccc}
        \toprule
        Model & NMSE $\downarrow$ & RMSE $\downarrow$ & SSIM $\uparrow$ \\
        \midrule
        RME-GAN & 0.0118 & 0.0307 & 0.9219 \\
        RadioUNet & 0.0089 & 0.0258 & 0.9410 \\
        UVM-Net & 0.0088 & 0.0301 & 0.9326 \\
        RadioDiff & 0.0057 & 0.0215 & 0.9536 \\ 
        \midrule
        \textbf{PhyRMDM (ours)} &\textbf{0.0047} &\textbf{0.0146} &\textbf{0.968} \\
        \bottomrule
    \end{tabular}
    \caption{Performance comparison of different models on the DRM setup in terms of NMSE, RMSE, and SSIM. PhyRMDM shows strong robustness in dynamic environments, achieving the lowest NMSE/RMSE and high SSIM, highlighting its ability to manage complexity from moving objects and varying propagation conditions.}
    \label{tab:setup2}
\end{table}

%% file: sec/tab/setup3.tex
\begin{table}[t]
    \centering
    \begin{tabular}{lcc}
        \toprule
        Model & NMSE $\downarrow$ & RMSE  $\downarrow$ \\
        \midrule
        
        AE & 0.2885 & 0.1238 \\
        Deep AE & 0.3152 & 0.1295 \\
       
        RadioUnet & 0.0042 & 0.0148 \\
        RME - GAN & 0.0036 & 0.0130 \\ \midrule
        \textbf{PhyRMDM (ours)} &\textbf{0.0022} & \textbf{0.0117} \\
        \bottomrule
    \end{tabular}
    \caption{Performance comparison of different models for Setup 3 using NMSE and RMSE. PhyRMDM achieves the lowest errors, showing its strength in handling unbalanced sample distributions and generating accurate radio maps under sparse, uneven observations.}
    \label{tab:setup3}
\end{table}

%% file: sec/tab/ablation.tex
\begin{table}[t]
    \centering
    \begin{tabular}{ccc|cc}
        \toprule
        $\mathcal{L}_\text{MSE}$  & $\mathcal{L}_\text{PINN}$ & $\mathcal{L}_\text{Reg}$ & NMSE $\downarrow$ & RMSE $\downarrow$  \\
        \midrule
        
        \checkmark & \checkmark & $\times$ & 0.0041 & 0.0155  \\
        $\times$ & \checkmark & \checkmark & 0.3716 & 0.2413  \\
        \checkmark & $\times$ & \checkmark & 0.0046 & 0.0182  \\
        \checkmark & \checkmark & \checkmark & 0.0031 & 0.0125  \\
        \bottomrule
    \end{tabular}
    \caption{Ablation results showing the effects of \$L\_\text{MSE}\$, \$L\_\text{PINN}\$, and \$L\_\text{Reg}\$ on NMSE and RMSE. \$L\_\text{MSE}\$ reduces prediction errors, \$L\_\text{PINN}\$ enforces physical consistency, and \$L\_\text{Reg}\$ improves stability and generalization. Combining all three yields the best performance, demonstrating their synergistic effect on accuracy and robustness.}
    \label{tab:ablation_study}
\end{table}

%% file: sec/conclusion.tex
\section{Conclusion}
\label{sec:conclusion}


We introduce the Physics-Informed Radio Map Diffusion Model (PhyRMDM), a framework for radio map reconstruction that integrates physics-based constraints with generative models. Directly constraining diffusion models is difficult due to complex radio environments and noise prediction. PhyRMDM tackles this with a dual U-Net design: one U-Net enforces physical alignment using an approximate Helmholtz equation, while the other refines the output through a parallel diffusion process. This approach produces accurate, robust radio maps even with sparse data. Experiments show PhyRMDM outperforms state-of-the-art methods across static, dynamic, and unbalanced scenarios, with notable improvements in NMSE, RMSE, and SSIM. While effective, the dual-U-Net structure is computationally heavy and sensitive to hyperparameters. Future work will focus on improving efficiency, adapting to dynamic conditions, and enabling real-time, large-scale applications with multi-source data and better interpretability for 5G/6G network optimization and spectrum management.

\newpage

%% file: main.bbl

\begin{thebibliography}{44}


\ifx \showCODEN    \undefined \def \showCODEN     #1{\unskip}     \fi
\ifx \showISBNx    \undefined \def \showISBNx     #1{\unskip}     \fi
\ifx \showISBNxiii \undefined \def \showISBNxiii  #1{\unskip}     \fi
\ifx \showISSN     \undefined \def \showISSN      #1{\unskip}     \fi
\ifx \showLCCN     \undefined \def \showLCCN      #1{\unskip}     \fi
\ifx \shownote     \undefined \def \shownote      #1{#1}          \fi
\ifx \showarticletitle \undefined \def \showarticletitle #1{#1}   \fi
\ifx \showURL      \undefined \def \showURL       {\relax}        \fi
\providecommand\bibfield[2]{#2}
\providecommand\bibinfo[2]{#2}
\providecommand\natexlab[1]{#1}
\providecommand\showeprint[2][]{arXiv:#2}

\bibitem[Alonazi et~al\mbox{.}(2017)]%
        {Interpolation1}
\bibfield{author}{\bibinfo{person}{Abdullah Alonazi}, \bibinfo{person}{Yi Ma}, {and} \bibinfo{person}{Rahim Tafazolli}.} \bibinfo{year}{2017}\natexlab{}.
\newblock \showarticletitle{Delaunay Triangulation Based Interpolation for Radio Map Construction with Reduced Calibration}. In \bibinfo{booktitle}{\emph{2017 9th IEEE-GCC Conference and Exhibition (GCCCE)}}. \bibinfo{pages}{1--9}.
\newblock
\href{https://doi.org/10.1109/IEEEGCC.2017.8448199}{doi:\nolinkurl{10.1109/IEEEGCC.2017.8448199}}


\bibitem[Audiffren and Bresciani(2022)]%
        {MBMF1l}
\bibfield{author}{\bibinfo{person}{Julien Audiffren} {and} \bibinfo{person}{Jean-Pierre Bresciani}.} \bibinfo{year}{2022}\natexlab{}.
\newblock \showarticletitle{Model based or model free? comparing adaptive methods for estimating thresholds in neuroscience}.
\newblock \bibinfo{journal}{\emph{Neural Computation}} \bibinfo{volume}{34}, \bibinfo{number}{2} (\bibinfo{year}{2022}), \bibinfo{pages}{338--359}.
\newblock


\bibitem[chen et~al\mbox{.}(2024)]%
        {sato}
\bibfield{author}{\bibinfo{person}{Wenshuo chen}, \bibinfo{person}{Hongru Xiao}, \bibinfo{person}{Erhang Zhang}, \bibinfo{person}{Lijie Hu}, \bibinfo{person}{Lei Wang}, \bibinfo{person}{Mengyuan Liu}, {and} \bibinfo{person}{Chen Chen}.} \bibinfo{year}{2024}\natexlab{}.
\newblock \showarticletitle{SATO: Stable Text-to-Motion Framework}. In \bibinfo{booktitle}{\emph{Proceedings of the 32nd ACM International Conference on Multimedia}} \emph{(\bibinfo{series}{MM ’24})}. \bibinfo{publisher}{ACM}, \bibinfo{pages}{6989–6997}.
\newblock
\href{https://doi.org/10.1145/3664647.3681034}{doi:\nolinkurl{10.1145/3664647.3681034}}


\bibitem[Croitoru et~al\mbox{.}(2023)]%
        {croitoru2023diffusion}
\bibfield{author}{\bibinfo{person}{Florinel-Alin Croitoru}, \bibinfo{person}{Vlad Hondru}, \bibinfo{person}{Radu~Tudor Ionescu}, {and} \bibinfo{person}{Mubarak Shah}.} \bibinfo{year}{2023}\natexlab{}.
\newblock \showarticletitle{Diffusion models in vision: A survey}.
\newblock \bibinfo{journal}{\emph{IEEE Transactions on Pattern Analysis and Machine Intelligence}} \bibinfo{volume}{45}, \bibinfo{number}{9} (\bibinfo{year}{2023}), \bibinfo{pages}{10850--10869}.
\newblock


\bibitem[Di~Matteo et~al\mbox{.}(2021)]%
        {GPSD}
\bibfield{author}{\bibinfo{person}{S Di~Matteo}, \bibinfo{person}{Nicholeen~M Viall}, {and} \bibinfo{person}{L Kepko}.} \bibinfo{year}{2021}\natexlab{}.
\newblock \showarticletitle{Power spectral density background estimate and signal detection via the multitaper method}.
\newblock \bibinfo{journal}{\emph{Journal of Geophysical Research: Space Physics}} \bibinfo{volume}{126}, \bibinfo{number}{2} (\bibinfo{year}{2021}), \bibinfo{pages}{e2020JA028748}.
\newblock


\bibitem[Eghtesad et~al\mbox{.}(2024)]%
        {eghtesad2024nn}
\bibfield{author}{\bibinfo{person}{Adnan Eghtesad}, \bibinfo{person}{Jingye Tan}, \bibinfo{person}{Jan~Niklas Fuhg}, {and} \bibinfo{person}{Nikolaos Bouklas}.} \bibinfo{year}{2024}\natexlab{}.
\newblock \showarticletitle{NN-EVP: A physics informed neural network-based elasto-viscoplastic framework for predictions of grain size-aware flow response}.
\newblock \bibinfo{journal}{\emph{International Journal of Plasticity}}  \bibinfo{volume}{181} (\bibinfo{year}{2024}), \bibinfo{pages}{104072}.
\newblock


\bibitem[Frei et~al\mbox{.}(2009)]%
        {frei2009prediction}
\bibfield{author}{\bibinfo{person}{Patrizia Frei}, \bibinfo{person}{Evelyn Mohler}, \bibinfo{person}{Alfred B{\"u}rgi}, \bibinfo{person}{J{\"u}rg Fr{\"o}hlich}, \bibinfo{person}{Georg Neubauer}, \bibinfo{person}{Charlotte Braun-Fahrl{\"a}nder}, \bibinfo{person}{Martin R{\"o}{\"o}sli}, {et~al\mbox{.}}} \bibinfo{year}{2009}\natexlab{}.
\newblock \showarticletitle{A prediction model for personal radio frequency electromagnetic field exposure}.
\newblock \bibinfo{journal}{\emph{Science of the total environment}} \bibinfo{volume}{408}, \bibinfo{number}{1} (\bibinfo{year}{2009}), \bibinfo{pages}{102--108}.
\newblock


\bibitem[Ho et~al\mbox{.}(2020)]%
        {ho2020denoisingdiffusionprobabilisticmodels}
\bibfield{author}{\bibinfo{person}{Jonathan Ho}, \bibinfo{person}{Ajay Jain}, {and} \bibinfo{person}{Pieter Abbeel}.} \bibinfo{year}{2020}\natexlab{}.
\newblock \bibinfo{title}{Denoising Diffusion Probabilistic Models}.
\newblock
\showeprint[arxiv]{2006.11239}~[cs.LG]
\urldef\tempurl%
\url{https://arxiv.org/abs/2006.11239}
\showURL{%
\tempurl}


\bibitem[Hu and McDaniel(2023)]%
        {hu2023applying}
\bibfield{author}{\bibinfo{person}{Beichao Hu} {and} \bibinfo{person}{Dwayne McDaniel}.} \bibinfo{year}{2023}\natexlab{}.
\newblock \showarticletitle{Applying physics-informed neural networks to solve Navier--Stokes equations for laminar flow around a particle}.
\newblock \bibinfo{journal}{\emph{Mathematical and Computational Applications}} \bibinfo{volume}{28}, \bibinfo{number}{5} (\bibinfo{year}{2023}), \bibinfo{pages}{102}.
\newblock


\bibitem[Hu et~al\mbox{.}(2024)]%
        {hu2024multidimensionalexplanationalignmentmedical}
\bibfield{author}{\bibinfo{person}{Lijie Hu}, \bibinfo{person}{Songning Lai}, \bibinfo{person}{Wenshuo Chen}, \bibinfo{person}{Hongru Xiao}, \bibinfo{person}{Hongbin Lin}, \bibinfo{person}{Lu Yu}, \bibinfo{person}{Jingfeng Zhang}, {and} \bibinfo{person}{Di Wang}.} \bibinfo{year}{2024}\natexlab{}.
\newblock \bibinfo{title}{Towards Multi-dimensional Explanation Alignment for Medical Classification}.
\newblock
\showeprint[arxiv]{2410.21494}~[cs.CV]
\urldef\tempurl%
\url{https://arxiv.org/abs/2410.21494}
\showURL{%
\tempurl}


\bibitem[Jakobus et~al\mbox{.}(2018)]%
        {jakobus2018recent}
\bibfield{author}{\bibinfo{person}{Ulrich Jakobus}, \bibinfo{person}{Andr{\'e}s~G Aguilar}, \bibinfo{person}{Gerd Woelfle}, \bibinfo{person}{Johann Van~Tonder}, \bibinfo{person}{Marianne Bingle}, \bibinfo{person}{Kitty Longtin}, {and} \bibinfo{person}{Martin Vogel}.} \bibinfo{year}{2018}\natexlab{}.
\newblock \showarticletitle{Recent advances of FEKO and WinProp}. In \bibinfo{booktitle}{\emph{2018 IEEE International Symposium on Antennas and Propagation \& USNC/URSI National Radio Science Meeting}}. IEEE, \bibinfo{pages}{409--410}.
\newblock


\bibitem[Jiang et~al\mbox{.}(2024)]%
        {jiang2024physics}
\bibfield{author}{\bibinfo{person}{Fenyu Jiang}, \bibinfo{person}{Tong Li}, \bibinfo{person}{Xingzai Lv}, \bibinfo{person}{Hua Rui}, {and} \bibinfo{person}{Depeng Jin}.} \bibinfo{year}{2024}\natexlab{}.
\newblock \showarticletitle{Physics-informed neural networks for path loss estimation by solving electromagnetic integral equations}.
\newblock \bibinfo{journal}{\emph{IEEE Transactions on Wireless Communications}} (\bibinfo{year}{2024}).
\newblock


\bibitem[Jornet et~al\mbox{.}(2024)]%
        {10579941}
\bibfield{author}{\bibinfo{person}{Josep~M. Jornet}, \bibinfo{person}{Vitaly Petrov}, \bibinfo{person}{Hua Wang}, \bibinfo{person}{Zoya Popović}, \bibinfo{person}{Dipankar Shakya}, \bibinfo{person}{Jose~V. Siles}, {and} \bibinfo{person}{Theodore~S. Rappaport}.} \bibinfo{year}{2024}\natexlab{}.
\newblock \showarticletitle{The Evolution of Applications, Hardware Design, and Channel Modeling for Terahertz (THz) Band Communications and Sensing: Ready for 6G?}
\newblock \bibinfo{journal}{\emph{Proc. IEEE}} (\bibinfo{year}{2024}), \bibinfo{pages}{1--32}.
\newblock
\href{https://doi.org/10.1109/JPROC.2024.3412828}{doi:\nolinkurl{10.1109/JPROC.2024.3412828}}


\bibitem[Jung et~al\mbox{.}(2011)]%
        {jung2011wi}
\bibfield{author}{\bibinfo{person}{Sukhoon Jung}, \bibinfo{person}{Choon-oh Lee}, {and} \bibinfo{person}{Dongsoo Han}.} \bibinfo{year}{2011}\natexlab{}.
\newblock \showarticletitle{Wi-Fi fingerprint-based approaches following log-distance path loss model for indoor positioning}. In \bibinfo{booktitle}{\emph{2011 IEEE MTT-S International Microwave Workshop Series on Intelligent Radio for Future Personal Terminals}}. IEEE, \bibinfo{pages}{1--2}.
\newblock


\bibitem[Juraev et~al\mbox{.}(2024)]%
        {juraev2024helmholtz}
\bibfield{author}{\bibinfo{person}{Davron~Aslonqulovich Juraev}, \bibinfo{person}{Praveen Agarwal}, \bibinfo{person}{Ebrahim~Eldesoky Elsayed}, {and} \bibinfo{person}{Nauryz Targyn}.} \bibinfo{year}{2024}\natexlab{}.
\newblock \showarticletitle{Helmholtz equations and their applications in solving physical problems}.
\newblock \bibinfo{journal}{\emph{Advanced Engineering Science}}  \bibinfo{volume}{4} (\bibinfo{year}{2024}), \bibinfo{pages}{54--64}.
\newblock


\bibitem[Kadir et~al\mbox{.}(2021)]%
        {9493470}
\bibfield{author}{\bibinfo{person}{Evizal~Abdul Kadir}, \bibinfo{person}{Raed Shubair}, \bibinfo{person}{Sharul~Kamal Abdul~Rahim}, \bibinfo{person}{Mohamed Himdi}, \bibinfo{person}{Muhammad~Ramlee Kamarudin}, {and} \bibinfo{person}{Sri~Listia Rosa}.} \bibinfo{year}{2021}\natexlab{}.
\newblock \showarticletitle{B5G and 6G: Next Generation Wireless Communications Technologies, Demand and Challenges}. In \bibinfo{booktitle}{\emph{2021 International Congress of Advanced Technology and Engineering (ICOTEN)}}. \bibinfo{pages}{1--6}.
\newblock
\href{https://doi.org/10.1109/ICOTEN52080.2021.9493470}{doi:\nolinkurl{10.1109/ICOTEN52080.2021.9493470}}


\bibitem[Kawar et~al\mbox{.}(2022)]%
        {kawar2022denoising}
\bibfield{author}{\bibinfo{person}{Bahjat Kawar}, \bibinfo{person}{Michael Elad}, \bibinfo{person}{Stefano Ermon}, {and} \bibinfo{person}{Jiaming Song}.} \bibinfo{year}{2022}\natexlab{}.
\newblock \showarticletitle{Denoising diffusion restoration models}.
\newblock \bibinfo{journal}{\emph{Advances in Neural Information Processing Systems}}  \bibinfo{volume}{35} (\bibinfo{year}{2022}), \bibinfo{pages}{23593--23606}.
\newblock


\bibitem[Keller and Borkowski(2019)]%
        {keller2019thin}
\bibfield{author}{\bibinfo{person}{Wolfgang Keller} {and} \bibinfo{person}{Andrzej Borkowski}.} \bibinfo{year}{2019}\natexlab{}.
\newblock \showarticletitle{Thin plate spline interpolation}.
\newblock \bibinfo{journal}{\emph{Journal of Geodesy}}  \bibinfo{volume}{93} (\bibinfo{year}{2019}), \bibinfo{pages}{1251--1269}.
\newblock


\bibitem[Lai et~al\mbox{.}(2024)]%
        {fts}
\bibfield{author}{\bibinfo{person}{Songning Lai}, \bibinfo{person}{Ninghui Feng}, \bibinfo{person}{Jiechao Gao}, \bibinfo{person}{Hao Wang}, \bibinfo{person}{Haochen Sui}, \bibinfo{person}{Xin Zou}, \bibinfo{person}{Jiayu Yang}, \bibinfo{person}{Wenshuo Chen}, \bibinfo{person}{Hang Zhao}, \bibinfo{person}{Xuming Hu}, {and} \bibinfo{person}{Yutao Yue}.} \bibinfo{year}{2024}\natexlab{}.
\newblock \bibinfo{title}{FTS: A Framework to Find a Faithful TimeSieve}.
\newblock
\showeprint[arxiv]{2405.19647}~[cs.LG]
\urldef\tempurl%
\url{https://arxiv.org/abs/2405.19647}
\showURL{%
\tempurl}


\bibitem[Lawal et~al\mbox{.}(2022)]%
        {lawal2022physics}
\bibfield{author}{\bibinfo{person}{Zaharaddeen~Karami Lawal}, \bibinfo{person}{Hayati Yassin}, \bibinfo{person}{Daphne Teck~Ching Lai}, {and} \bibinfo{person}{Azam Che~Idris}.} \bibinfo{year}{2022}\natexlab{}.
\newblock \showarticletitle{Physics-informed neural network (PINN) evolution and beyond: A systematic literature review and bibliometric analysis}.
\newblock \bibinfo{journal}{\emph{Big Data and Cognitive Computing}} \bibinfo{volume}{6}, \bibinfo{number}{4} (\bibinfo{year}{2022}), \bibinfo{pages}{140}.
\newblock


\bibitem[Li et~al\mbox{.}(2020)]%
        {LI20203394}
\bibfield{author}{\bibinfo{person}{Peng Li}, \bibinfo{person}{Xiaoping Lu}, {and} \bibinfo{person}{Song-Ping Zhu}.} \bibinfo{year}{2020}\natexlab{}.
\newblock \showarticletitle{Pricing weather derivatives with the market price of risk extracted from the utility indifference valuation}.
\newblock \bibinfo{journal}{\emph{Computers \& Mathematics with Applications}} \bibinfo{volume}{79}, \bibinfo{number}{12} (\bibinfo{year}{2020}), \bibinfo{pages}{3394--3409}.
\newblock
\showISSN{0898-1221}
\href{https://doi.org/10.1016/j.camwa.2020.02.007}{doi:\nolinkurl{10.1016/j.camwa.2020.02.007}}


\bibitem[Li et~al\mbox{.}(2024)]%
        {RadioGAT}
\bibfield{author}{\bibinfo{person}{Xiaojie Li}, \bibinfo{person}{Songyang Zhang}, \bibinfo{person}{Hang Li}, \bibinfo{person}{Xiaoyang Li}, \bibinfo{person}{Lexi Xu}, \bibinfo{person}{Haigao Xu}, \bibinfo{person}{Hui Mei}, \bibinfo{person}{Guangxu Zhu}, \bibinfo{person}{Nan Qi}, {and} \bibinfo{person}{Ming Xiao}.} \bibinfo{year}{2024}\natexlab{}.
\newblock \showarticletitle{RadioGAT: A Joint Model-Based and Data-Driven Framework for Multi-Band Radiomap Reconstruction via Graph Attention Networks}.
\newblock \bibinfo{journal}{\emph{IEEE Transactions on Wireless Communications}} \bibinfo{volume}{23}, \bibinfo{number}{11} (\bibinfo{date}{Nov.} \bibinfo{year}{2024}), \bibinfo{pages}{17777–17792}.
\newblock
\showISSN{1558-2248}
\href{https://doi.org/10.1109/twc.2024.3457157}{doi:\nolinkurl{10.1109/twc.2024.3457157}}


\bibitem[Luo et~al\mbox{.}(2024)]%
        {luo2024rm}
\bibfield{author}{\bibinfo{person}{Xuanhao Luo}, \bibinfo{person}{L Zhizhen}, \bibinfo{person}{Zhiyuan Peng}, \bibinfo{person}{X Dongkuan}, {and} \bibinfo{person}{Yuchen Liu}.} \bibinfo{year}{2024}\natexlab{}.
\newblock \showarticletitle{Rm-gen: Conditional diffusion model-based radio map generation for wireless networks}. In \bibinfo{booktitle}{\emph{2024 IFIP Networking Conference (IFIP Networking)}}. IEEE, \bibinfo{pages}{543--548}.
\newblock


\bibitem[L’heureux et~al\mbox{.}(2017)]%
        {l2017machine}
\bibfield{author}{\bibinfo{person}{Alexandra L’heureux}, \bibinfo{person}{Katarina Grolinger}, \bibinfo{person}{Hany~F Elyamany}, {and} \bibinfo{person}{Miriam~AM Capretz}.} \bibinfo{year}{2017}\natexlab{}.
\newblock \showarticletitle{Machine learning with big data: Challenges and approaches}.
\newblock \bibinfo{journal}{\emph{Ieee Access}}  \bibinfo{volume}{5} (\bibinfo{year}{2017}), \bibinfo{pages}{7776--7797}.
\newblock


\bibitem[Mantiply et~al\mbox{.}(1997)]%
        {RF-filed-measurement}
\bibfield{author}{\bibinfo{person}{Edwin~D Mantiply}, \bibinfo{person}{Kenneth~R Pohl}, \bibinfo{person}{Samuel~W Poppell}, {and} \bibinfo{person}{Julia~A Murphy}.} \bibinfo{year}{1997}\natexlab{}.
\newblock \showarticletitle{Summary of measured radiofrequency electric and magnetic fields (10 kHz to 30 GHz) in the general and work environment}.
\newblock \bibinfo{journal}{\emph{Bioelectromagnetics: Journal of the Bioelectromagnetics Society, The Society for Physical Regulation in Biology and Medicine, The European Bioelectromagnetics Association}} \bibinfo{volume}{18}, \bibinfo{number}{8} (\bibinfo{year}{1997}), \bibinfo{pages}{563--577}.
\newblock


\bibitem[Monaco and Apiletti(2023)]%
        {monaco2023training}
\bibfield{author}{\bibinfo{person}{Simone Monaco} {and} \bibinfo{person}{Daniele Apiletti}.} \bibinfo{year}{2023}\natexlab{}.
\newblock \showarticletitle{Training physics-informed neural networks: One learning to rule them all?}
\newblock \bibinfo{journal}{\emph{Results in Engineering}}  \bibinfo{volume}{18} (\bibinfo{year}{2023}), \bibinfo{pages}{101023}.
\newblock


\bibitem[Nabian et~al\mbox{.}(2021)]%
        {nabian2021efficient}
\bibfield{author}{\bibinfo{person}{Mohammad~Amin Nabian}, \bibinfo{person}{Rini~Jasmine Gladstone}, {and} \bibinfo{person}{Hadi Meidani}.} \bibinfo{year}{2021}\natexlab{}.
\newblock \showarticletitle{Efficient training of physics-informed neural networks via importance sampling}.
\newblock \bibinfo{journal}{\emph{Computer-Aided Civil and Infrastructure Engineering}} \bibinfo{volume}{36}, \bibinfo{number}{8} (\bibinfo{year}{2021}), \bibinfo{pages}{962--977}.
\newblock


\bibitem[Nichol and Dhariwal(2021)]%
        {diff1-nichol2021improved}
\bibfield{author}{\bibinfo{person}{Alexander~Quinn Nichol} {and} \bibinfo{person}{Prafulla Dhariwal}.} \bibinfo{year}{2021}\natexlab{}.
\newblock \showarticletitle{Improved denoising diffusion probabilistic models}. In \bibinfo{booktitle}{\emph{International conference on machine learning}}. PMLR, \bibinfo{pages}{8162--8171}.
\newblock


\bibitem[Ning et~al\mbox{.}(2024)]%
        {ning2024dctdiffintriguingpropertiesimage}
\bibfield{author}{\bibinfo{person}{Mang Ning}, \bibinfo{person}{Mingxiao Li}, \bibinfo{person}{Jianlin Su}, \bibinfo{person}{Haozhe Jia}, \bibinfo{person}{Lanmiao Liu}, \bibinfo{person}{Martin Beneš}, \bibinfo{person}{Albert~Ali Salah}, {and} \bibinfo{person}{Itir~Onal Ertugrul}.} \bibinfo{year}{2024}\natexlab{}.
\newblock \bibinfo{title}{DCTdiff: Intriguing Properties of Image Generative Modeling in the DCT Space}.
\newblock
\showeprint[arxiv]{2412.15032}~[cs.CV]
\urldef\tempurl%
\url{https://arxiv.org/abs/2412.15032}
\showURL{%
\tempurl}


\bibitem[Raissi et~al\mbox{.}(2019a)]%
        {RAISSI2019686}
\bibfield{author}{\bibinfo{person}{M. Raissi}, \bibinfo{person}{P. Perdikaris}, {and} \bibinfo{person}{G.E. Karniadakis}.} \bibinfo{year}{2019}\natexlab{a}.
\newblock \showarticletitle{Physics-informed neural networks: A deep learning framework for solving forward and inverse problems involving nonlinear partial differential equations}.
\newblock \bibinfo{journal}{\emph{J. Comput. Phys.}}  \bibinfo{volume}{378} (\bibinfo{year}{2019}), \bibinfo{pages}{686--707}.
\newblock
\showISSN{0021-9991}
\href{https://doi.org/10.1016/j.jcp.2018.10.045}{doi:\nolinkurl{10.1016/j.jcp.2018.10.045}}


\bibitem[Raissi et~al\mbox{.}(2019b)]%
        {raissi2019physics}
\bibfield{author}{\bibinfo{person}{Maziar Raissi}, \bibinfo{person}{Paris Perdikaris}, {and} \bibinfo{person}{George~E Karniadakis}.} \bibinfo{year}{2019}\natexlab{b}.
\newblock \showarticletitle{Physics-informed neural networks: A deep learning framework for solving forward and inverse problems involving nonlinear partial differential equations}.
\newblock \bibinfo{journal}{\emph{Journal of Computational physics}}  \bibinfo{volume}{378} (\bibinfo{year}{2019}), \bibinfo{pages}{686--707}.
\newblock


\bibitem[Ramesh et~al\mbox{.}(2021)]%
        {ramesh2021zeroshottexttoimagegeneration}
\bibfield{author}{\bibinfo{person}{Aditya Ramesh}, \bibinfo{person}{Mikhail Pavlov}, \bibinfo{person}{Gabriel Goh}, \bibinfo{person}{Scott Gray}, \bibinfo{person}{Chelsea Voss}, \bibinfo{person}{Alec Radford}, \bibinfo{person}{Mark Chen}, {and} \bibinfo{person}{Ilya Sutskever}.} \bibinfo{year}{2021}\natexlab{}.
\newblock \bibinfo{title}{Zero-Shot Text-to-Image Generation}.
\newblock
\showeprint[arxiv]{2102.12092}~[cs.CV]
\urldef\tempurl%
\url{https://arxiv.org/abs/2102.12092}
\showURL{%
\tempurl}


\bibitem[Rombach et~al\mbox{.}(2022)]%
        {rombach2022highresolutionimagesynthesislatent}
\bibfield{author}{\bibinfo{person}{Robin Rombach}, \bibinfo{person}{Andreas Blattmann}, \bibinfo{person}{Dominik Lorenz}, \bibinfo{person}{Patrick Esser}, {and} \bibinfo{person}{Björn Ommer}.} \bibinfo{year}{2022}\natexlab{}.
\newblock \bibinfo{title}{High-Resolution Image Synthesis with Latent Diffusion Models}.
\newblock
\showeprint[arxiv]{2112.10752}~[cs.CV]
\urldef\tempurl%
\url{https://arxiv.org/abs/2112.10752}
\showURL{%
\tempurl}


\bibitem[Ronneberger et~al\mbox{.}(2015)]%
        {ronneberger2015u}
\bibfield{author}{\bibinfo{person}{Olaf Ronneberger}, \bibinfo{person}{Philipp Fischer}, {and} \bibinfo{person}{Thomas Brox}.} \bibinfo{year}{2015}\natexlab{}.
\newblock \showarticletitle{U-net: Convolutional networks for biomedical image segmentation}. In \bibinfo{booktitle}{\emph{Medical image computing and computer-assisted intervention--MICCAI 2015: 18th international conference, Munich, Germany, October 5-9, 2015, proceedings, part III 18}}. Springer, \bibinfo{pages}{234--241}.
\newblock


\bibitem[Sizun(2005)]%
        {sizun2005introduction}
\bibfield{author}{\bibinfo{person}{Herv{\'e} Sizun}.} \bibinfo{year}{2005}\natexlab{}.
\newblock \showarticletitle{Introduction to the Propagation of Radio Waves}.
\newblock \bibinfo{journal}{\emph{Radio Wave Propagation for Telecommunication Applications}} (\bibinfo{year}{2005}), \bibinfo{pages}{1--12}.
\newblock


\bibitem[Sloan et~al\mbox{.}(2012)]%
        {sloan2012partial}
\bibfield{author}{\bibinfo{person}{D Sloan}, \bibinfo{person}{S Vandewalle}, {and} \bibinfo{person}{E S{\"u}li}.} \bibinfo{year}{2012}\natexlab{}.
\newblock \showarticletitle{Partial differential equations}.
\newblock  (\bibinfo{year}{2012}).
\newblock


\bibitem[Song et~al\mbox{.}(2020)]%
        {song2020denoising}
\bibfield{author}{\bibinfo{person}{Jiaming Song}, \bibinfo{person}{Chenlin Meng}, {and} \bibinfo{person}{Stefano Ermon}.} \bibinfo{year}{2020}\natexlab{}.
\newblock \showarticletitle{Denoising diffusion implicit models}.
\newblock \bibinfo{journal}{\emph{arXiv preprint arXiv:2010.02502}} (\bibinfo{year}{2020}).
\newblock


\bibitem[Song et~al\mbox{.}(2022)]%
        {song2022denoisingdiffusionimplicitmodels}
\bibfield{author}{\bibinfo{person}{Jiaming Song}, \bibinfo{person}{Chenlin Meng}, {and} \bibinfo{person}{Stefano Ermon}.} \bibinfo{year}{2022}\natexlab{}.
\newblock \bibinfo{title}{Denoising Diffusion Implicit Models}.
\newblock
\showeprint[arxiv]{2010.02502}~[cs.LG]
\urldef\tempurl%
\url{https://arxiv.org/abs/2010.02502}
\showURL{%
\tempurl}


\bibitem[Wang et~al\mbox{.}(2024)]%
        {wang2024radiodiff}
\bibfield{author}{\bibinfo{person}{Xiucheng Wang}, \bibinfo{person}{Keda Tao}, \bibinfo{person}{Nan Cheng}, \bibinfo{person}{Zhisheng Yin}, \bibinfo{person}{Zan Li}, \bibinfo{person}{Yuan Zhang}, {and} \bibinfo{person}{Xuemin Shen}.} \bibinfo{year}{2024}\natexlab{}.
\newblock \showarticletitle{RadioDiff: An Effective Generative Diffusion Model for Sampling-Free Dynamic Radio Map Construction}.
\newblock \bibinfo{journal}{\emph{IEEE Transactions on Cognitive Communications and Networking}} (\bibinfo{year}{2024}).
\newblock


\bibitem[Webbink(1977)]%
        {Value_of_Frequency_Spectrum}
\bibfield{author}{\bibinfo{person}{Douglas~W. Webbink}.} \bibinfo{year}{1977}\natexlab{}.
\newblock \showarticletitle{The Value of the Frequency Spectrum Allocated to Specific Uses}.
\newblock \bibinfo{journal}{\emph{IEEE Transactions on Electromagnetic Compatibility}} \bibinfo{volume}{EMC-19}, \bibinfo{number}{3} (\bibinfo{year}{1977}), \bibinfo{pages}{343--351}.
\newblock
\href{https://doi.org/10.1109/TEMC.1977.303607}{doi:\nolinkurl{10.1109/TEMC.1977.303607}}


\bibitem[Yang et~al\mbox{.}(2023)]%
        {yang2023diffusion}
\bibfield{author}{\bibinfo{person}{Ling Yang}, \bibinfo{person}{Zhilong Zhang}, \bibinfo{person}{Yang Song}, \bibinfo{person}{Shenda Hong}, \bibinfo{person}{Runsheng Xu}, \bibinfo{person}{Yue Zhao}, \bibinfo{person}{Wentao Zhang}, \bibinfo{person}{Bin Cui}, {and} \bibinfo{person}{Ming-Hsuan Yang}.} \bibinfo{year}{2023}\natexlab{}.
\newblock \showarticletitle{Diffusion models: A comprehensive survey of methods and applications}.
\newblock \bibinfo{journal}{\emph{Comput. Surveys}} \bibinfo{volume}{56}, \bibinfo{number}{4} (\bibinfo{year}{2023}), \bibinfo{pages}{1--39}.
\newblock


\bibitem[Yapar et~al\mbox{.}(2022)]%
        {rms}
\bibfield{author}{\bibinfo{person}{Cagkan Yapar}, \bibinfo{person}{Ron Levie}, \bibinfo{person}{Gitta Kutyniok}, {and} \bibinfo{person}{Giuseppe Caire}.} \bibinfo{year}{2022}\natexlab{}.
\newblock \bibinfo{title}{Dataset of Pathloss and ToA Radio Maps with Localization Application}.
\newblock
\href{https://doi.org/10.21227/0gtx-6v30}{doi:\nolinkurl{10.21227/0gtx-6v30}}


\bibitem[Yurtsever et~al\mbox{.}(2020)]%
        {9046805}
\bibfield{author}{\bibinfo{person}{Ekim Yurtsever}, \bibinfo{person}{Jacob Lambert}, \bibinfo{person}{Alexander Carballo}, {and} \bibinfo{person}{Kazuya Takeda}.} \bibinfo{year}{2020}\natexlab{}.
\newblock \showarticletitle{A Survey of Autonomous Driving: Common Practices and Emerging Technologies}.
\newblock \bibinfo{journal}{\emph{IEEE Access}}  \bibinfo{volume}{8} (\bibinfo{year}{2020}), \bibinfo{pages}{58443--58469}.
\newblock
\href{https://doi.org/10.1109/ACCESS.2020.2983149}{doi:\nolinkurl{10.1109/ACCESS.2020.2983149}}


\bibitem[Zhang et~al\mbox{.}(2023)]%
        {REM-GAN}
\bibfield{author}{\bibinfo{person}{Songyang Zhang}, \bibinfo{person}{Achintha Wijesinghe}, {and} \bibinfo{person}{Zhi Ding}.} \bibinfo{year}{2023}\natexlab{}.
\newblock \showarticletitle{RME-GAN: A Learning Framework for Radio Map Estimation Based on Conditional Generative Adversarial Network}.
\newblock \bibinfo{journal}{\emph{IEEE Internet of Things Journal}} \bibinfo{volume}{10}, \bibinfo{number}{20} (\bibinfo{year}{2023}), \bibinfo{pages}{18016--18027}.
\newblock
\href{https://doi.org/10.1109/JIOT.2023.3278235}{doi:\nolinkurl{10.1109/JIOT.2023.3278235}}


\end{thebibliography}
